\documentclass[twoside]{article} 
\usepackage{aistats2017}
\usepackage{times}

\usepackage[utf8]{inputenc} 
\usepackage[T1]{fontenc}    
\usepackage{hyperref}       
\usepackage{url}            
\usepackage{booktabs}       
\usepackage{amsfonts}       
\usepackage{nicefrac}       
\usepackage{microtype}      
\usepackage{Definitions}
\usepackage{adjustbox}
\usepackage{float}
\usepackage{subfigure} 
\usepackage{algorithm,algorithmic}
\usepackage{color}
\usepackage{xspace}
\usepackage{enumitem}

\usepackage{color}

\begin{document}

\twocolumn[

\aistatstitle{Data-Driven Threshold Machine: Scan Statistics, Change-Point Detection, and Extreme Bandits}

\aistatsauthor{Shuang Li \And Yao Xie \And Le Song }

\aistatsaddress{School of Industrial \\and Systems Engineering \\Georgia Tech\\Email: sli370@gatech.edu \And School of Industrial \\and Systems Engineering \\Georgia Tech\\Email: yao.xie@isye.gatech.edu \ \And School of Computer Science \\and Engineering\\Georgia Tech\\Email: lsong@cc.gatech.edu} ]

\begin{abstract}

We present a novel distribution-free approach, the data-driven threshold machine (DTM), for 
a fundamental problem at the core of many learning tasks: choose a threshold for a given pre-specified level that bounds the tail probability of the maximum of a (possibly dependent but stationary) random sequence. We do not assume data distribution, but rather relying on the asymptotic distribution of extremal values, and reduce the problem to estimate three parameters of the extreme value distributions and the extremal index. We specially take care of data dependence via estimating extremal index since in many settings, such as scan statistics, change-point detection, and extreme bandits, where dependence in the sequence of statistics can be significant. Key features of our DTM also include robustness and the computational efficiency, and it only requires one sample path to form a reliable estimate of the threshold, in contrast to the Monte Carlo sampling approach which 
requires drawing a large number of sample paths. We demonstrate the good performance of DTM via numerical examples in various dependent settings.

\end{abstract}

\vspace{-0.1in}

\vspace{-4mm}
\section{Introduction}
\vspace{-3mm}

\setlength{\abovedisplayskip}{4pt}
\setlength{\abovedisplayshortskip}{1pt}
\setlength{\belowdisplayskip}{4pt}
\setlength{\belowdisplayshortskip}{1pt}
\setlength{\jot}{3pt}

Selecting threshold is a key step in many machine learning tasks, such as anomaly detection by scan statistics \cite{Scan09}, sequential change-point detection \cite{annals2013}, and extreme $K$-arm bandit \cite{extremeBandits14}. Broadly speaking, determining threshold is the central problem for statistical hypothesis testing and estimating confidence intervals. The goal of setting the threshold include controlling the $p$-value or the significance level, controlling the false-alarm rate, or establishing the upper or lower confidence bounds for the max $K$-arm bandits. This goal can usually be cast into setting a threshold $x$ such that the maximum of a random sequence $S_1, S_2, \ldots, S_n$, which typically corresponds to statistics or  outputs of a learning algorithm, is less than the threshold with a pre-specified level $\alpha$, i.e., 
\begin{equation}
\mathbb{P}\cbr{\max_{t=1,\ldots n} S_t > x} \leqslant \alpha, \label{key_problem}
\end{equation}
under the assumed distribution in the hypothesis setting etc. These $S_t$ are dependent in many settings. For instance, in scan statistics, there are $n$ scanning regions, $S_t$ corresponds to a statistic formed for each region. An anomaly is detected if any of the regional statistics exceeds the threshold, and $\alpha$ is the significance level.

Despite its importance, setting threshold remains one of the most challenging parts in designing a learning algorithm. This is commonly done by Monte Carlo simulations and bootstrapping, which requires repeating a large number of experiments to generate sequences either from the assumed distribution or by bootstrapping the original sequence; this can be computationally extensive. Since $\alpha$ is usually set to a small number (for the algorithm to be reliable), this means that we have to estimate a small probability. To obtain a high precision, a large number of repetitions are needed. What exacerbates this is that in many settings generating samples are not easy. For instance, the assumed distribution can be difficult to draw samples, and it is common to use the computationally extensive Markov-Chain Monte Carlo techniques. In the learning setting, this can mean to run the algorithms many times, and running the  algorithm even once (such as deep-learning) over a large-scale reference dataset even once can 
be time-consuming.

In other cases, analytical approximations are found to relate the tail probability to the threshold (e.g., ARL approximation in the sequential change-point detection setting \cite{annals2013}). However, these results typically make strong parametric assumptions on the data  to make the problem tractable. In practice it is hard to model the distribution for the sequence precisely, being the output of a learning algorithm, and the distribution may vary from one dataset to the next. Moreover, the random sequence has non-negligible dependence, while theoretical approximations are usually developed for~\iid~samples. For instance, in scan statistics \cite{Scan09} for anomaly detection, a detection statistic is formed for each local scanning region, and the statistics for overlapping scanning regions are correlated since they are computed over common data. In sequential hypothesis testing and change-point detection, given a streaming data sequence, each time we form a statistic over a past sliding window to 
detect a change. Due to overlapping of the sliding window, the detection statistics at each time are highly correlated. In the bandit setting, the empirical rewards may be estimated from a sliding window which again results in dependence. Without taking into account the dependence, threshold choice is usually inaccurate and cannot meet the targeted level.

Therefore, there is a substantial need for developing a unifying framework for designing threshold addressing the above issues. The proposed approach should be computationally efficient, distribution free, accurate, and robust to the dependence inherent to the sequence of statistics.  

{\bf Our contributions:} In this paper, we present a novel distribution-free approach to choosing a threshold for a broad class of algorithms given a pre-specified level, which we call the data-driven threshold machine (DTM). DTM takes a (possibly dependent but stationary) sequence $\{S_1, \ldots, S_n\}$ generated by a learning algorithm, a pre-specified level $\alpha$, and returns a threshold $x$ such that $\mathbb{P}\{\max_{i=1}^n S_i > x\} \leqslant \alpha$ (illustrated in Figure~\ref{fig:dtm} below). 

\begin{figure}[h!]
\label{fig:dtm}
\vspace{-3mm}
\centering 
\includegraphics[width = 0.9\linewidth]{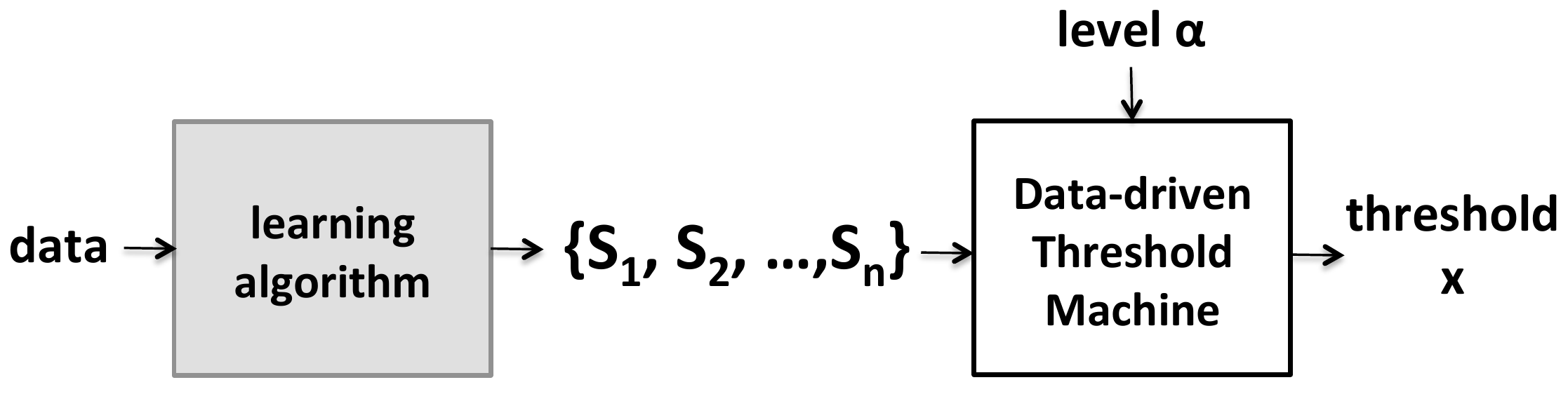}
\vspace{-3mm}
\caption{Diagram of data-driven machine (DTM). The input of DTM will be a (possibly dependent but stationary) sequence generated by some learning algorithm and a pre-specified small number, level $\alpha$. The output will be a threshold $x$ such as $\mathbb{P}\{\max_{i=1}^n S_i > x\} \leqslant \alpha$. }
\end{figure}

We make a connection between the threshold design and the extreme value theory (EVT) since the threshold design can be cast into a problem of determining the tail of the extreme over a random sequence. The classic literature of EVT \cite{Leadbetter83, Hsing88} has been focused on developing the limiting distribution of the extreme and the estimation parts based on theory are written obscurely in various scatted places. In the past, EVT has been largely used for domains such as finance \cite{EVTRisk06,Rocco12} and environmental science \cite{Smith89}. 
In this paper, we focus on estimation by using the forms of the limiting distributions from the classic references \cite{Leadbetter83, Hsing88}, but also take advantage of the advances in EVT \cite{Ferro03,Suveges07} to handle dependent sequence via estimating the extremal index, which is quite important to settings such as scan statistics, online change-point detection, and extreme bandits, where dependent between the sequence of statistics can be significant. Moreover, EVT directly focuses on the tail of distribution, thus avoiding a drawback of using statistical approximation, 
whose estimates are somehow biased by the central part of the distribution. In principle, EVT-based estimates of threshold can be more precise.

In a nutshell, our approach is to relate threshold to the tail probability of an arbitrary dependence sequence. We leverage the forms of the limiting distribution functions to parameterize the tail probability using four parameters including the extremal index, which explicitly captures the dependence. DMT is a three-stage method. In Stage I, we bootstrap from the original sequence of samples to generate an~\iid~sequence with the same marginal distribution. In Stage II, we estimate the parameters of the extreme value distributions, using the heights of the exceedance points in this~\iid~sequence given a pre-specified threshold. In Stage III, we estimate the extremal index using the inter-exceedance times of the original sequence. 
To summarize, the features of DMT include
\begin{itemize}[leftmargin=*,nosep,nolistsep]
 
\item DMT is {\it distribution-free} without making any parametric assumption about the sequence. To get around assuming the parametric distribution for data, we use the asymptotic distribution of the maximum of a sequence. The ideas leverage the powerful extreme type theorem, which states that the limiting distribution of the maxima will be one of the three distributions, Weibull, Gumbel or Fr\'{e}chet law. Hence, this reduces the task of estimating the tail probability to a much simpler one of estimating three parameters of the extreme value distributions. The asymptotic kicks in with a moderate sample size \cite{Leadbetter83}. The samples are utilized to estimate these parameters as well as the {\it extremal index} described below, via a ``Poisson process'' trick: when the threshold value high, the exceedance events are rare and can be well modeled as a Poisson process.

\item DTM is {\it robust to dependence} of the sequence. It can obtain accurate threshold even when the sequence is dependent and works well as long as the sequence does not have infinite memory. 

\item DTM is {\it computationally efficient}. Since it only takes the original sequence, without performing any Monte Carlo simulation. The main computation involves maximum likelihood estimation of fours parameters where many standard optimization procedures can be employed. 
\end{itemize}

\vspace{-4mm}
\subsection{Closely related work}
\vspace{-3mm}

Choosing threshold using EVT has been studied in 
\cite{threshold10}; however, they assume~\iid~samples, which cannot be applied to the settings we consider here such as scan-statistic, change-point detection since the dependence in the sequence of statistics is very significant. In other settings, EVT has been used to understand the theoretical basis of machine learning algorithms: recognition score analysis \cite{ECCV10,PAMI11}, for novelty detection \cite{Novelty09}, and for satellite image analysis \cite{Shao13}. 

\vspace{-4mm}
\subsection{Motivating examples}
\vspace{-3mm}

{\bf Scan statistics \cite{Scan09}.} There are $n$ scanning regions,  $S_t$ corresponds to a statistic formed for each region; an anomaly is detected when any of the regions has statistic exceeds the threshold. The probability is over the null distribution assuming there is no anomaly, and $\alpha$ is pre-specified type-I error or significance level. Thus, the definition of significance level is (\ref{key_problem}). 



{\bf Online change-point detection \cite{Siegmund1985}.} Given a sequence of mutually independent data $\{x_1, x_2, \dots\}$, there may be a change-point such that the distribution of the data changes. Our goal is detect such a change as quickly as possible after it occurs. The well-known CUSUM procedure uses a log-likelihood ratio statistic $S_t:= \max_{k<t} \sum_{i=k+1}^t \ell(x_i)$ for each time $t$, where $\ell(x_i)$ is the log-likelihood for each individual sample and the maximizing over $k$ corresponds to searching for the unknown change-point location. The detection procedure is a stopping time $T = \inf\{t: S_t > x\}$. To control the false-alarm-rate, one will specify the so-called average-run-length (ARL) so that $\mathbb{E}_0(T) \leq {\rm ARL}$. It can be shown that $T$ is asymptotically exponential when $x$ is large \cite{SiegmundVenkatraman1995}, and hence the ARL requirement can be translated into $\mathbb{P}_0\{\max_{1\leqslant i \leqslant n} S_i > x\} = 1-e^{-n/{\rm ARL}}$. The sequential 
change-point detection can be viewed as a special case of the sequential likelihood ratio test (SPRT) \cite{Siegmund1985}, in which similar relations between the threshold and the specified levels occur. 

{\bf Extreme bandits.} 
The extreme bandits \cite{extremeBandits14}, also known as the max-$K$ bandit in \cite{kArm05}, models a scenario in outlier detection, security, and medicine. It considers the following learning setting. The learner chooses one arm each time and then receives only the sample for that arm. For each pull, the $k$th arm generates a reward following a distribution $f_k$ with unknown parameters. Let $S_{k, t}$ be the estimate for the true reward. The estimate for $S_{k, t}$, if using sliding window, will have non-negligible dependence. The performance of a learner is evaluated by the {\it most extreme value} she has found. In this setting, to use the classic upper confidence bound rule (see, e.g., \cite{UCB2014}), one has to find $\mathbb{P}_k\{\max_{1\leqslant i\leqslant t}S_{k, i} > x\} < \alpha$ for each arm $k$ for a pre-specified confidence level $1-\alpha$. 


\vspace{-4mm}
\section{Data-Driven Threshold Machine}
\vspace{-3mm}

Given a sequence of (possibly dependent but stationary) observations of length $n$
$$\mathbb{S} = \{S_1, S_2, \ldots, S_n\}$$ 
generated as the output of a learning algorithm, our data-driven threshold machine (DTM) returns the threshold $x$ for a certain target level $\alpha$ in three steps:  
\begin{itemize}[leftmargin=*,nosep,nolistsep]
  \item[\bf I] The algorithm first bootstraps (or samples with replacement) from the original sequence $\mathbb{S}$ to generate a new~\iid~sequence 
  $$\mathbb{S}^* = \{S_1^*, S_2^*, \dots, S_n^*\}.$$
  Due to the sampling technique, the new sequence preserves the marginal distribution but breaks the local dependence in the original stationary sequence. 
  \item[\bf II] The algorithm selects {\it exceedant} samples which are greater than a large pre-set cutoff value $u$ from $\mathbb{S}^*$. The index and height of these exceedant sample will follow a marked Poisson process approximately, and we use them to estimate the (location, scale, and type) parameters of the extreme value distribution. (Illustrated in Fig. \ref{demo1}).
  \begin{figure}[H]
  \vspace{-3mm}
    \centering               
    \includegraphics[width=0.8 \linewidth]{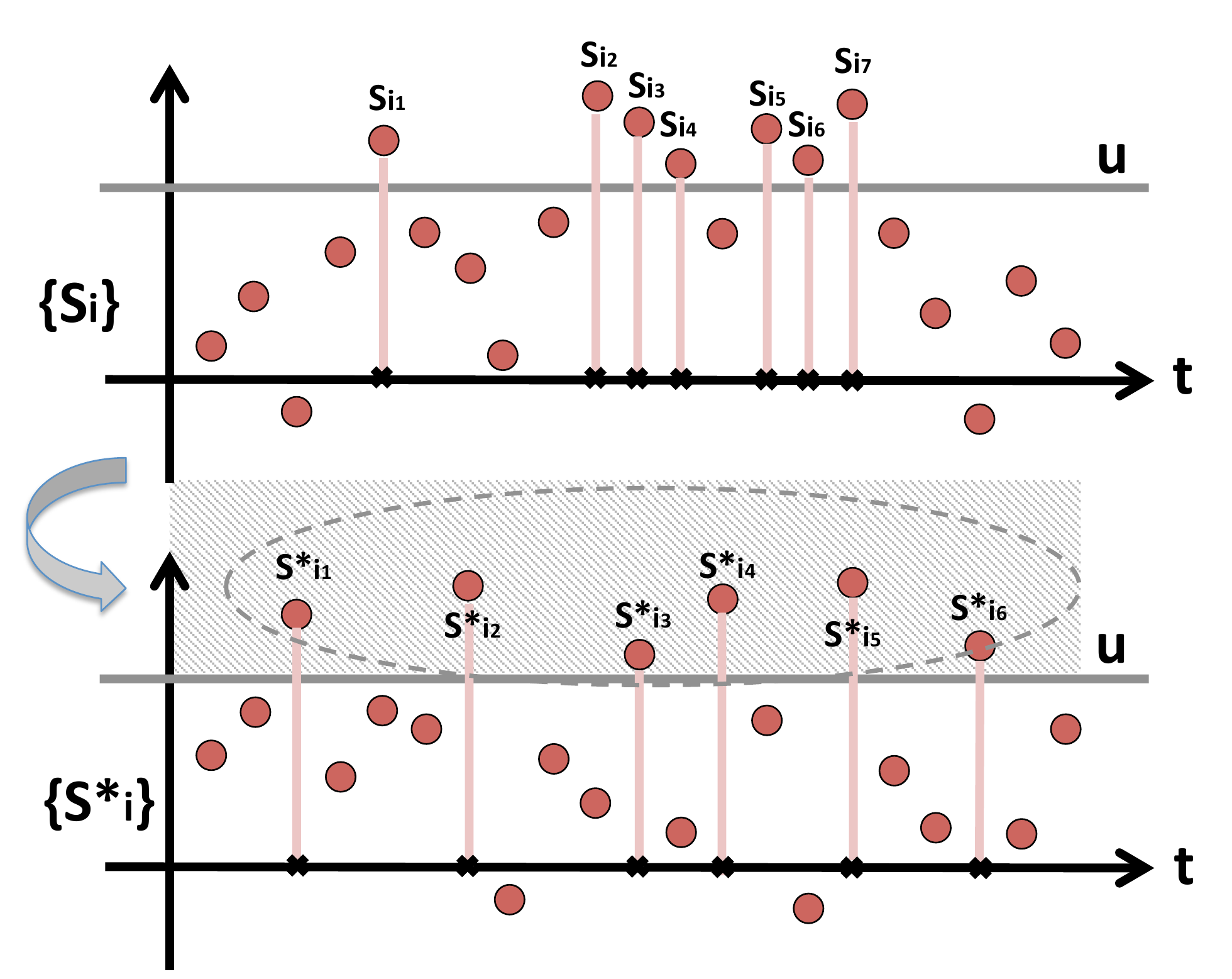}
    \vspace{-3mm}
    \caption{Stage I and II: Bootstrapping to obtain $\mathbb{S}^*$, apply cutoff $u$ to $\mathbb{S}^*$ to obtain a marked Poisson process, and estimate the (location, scale, and type) parameters of the extreme value distribution.}    
    \label{demo1}
  \vspace{-3mm}
  \end{figure}  
  \item[\bf III] The algorithm returns to the original sequence $\mathbb{S}$, and apply the same pre-set cutoff value $u$. This is based on the profound theory that the threshold exceeding events converges in distribution to a compound Poisson process~\cite{Leadbetter83}. Then the algorithm estimates the extremal index to capture inter-dependence between samples, using the temporal intervals between adjacent exceedant points. (Illustrated in Fig. \ref{demo2}).
  \begin{figure}[H]
  \vspace{-3mm}
    \centering               
    \includegraphics[width=0.8 \linewidth]{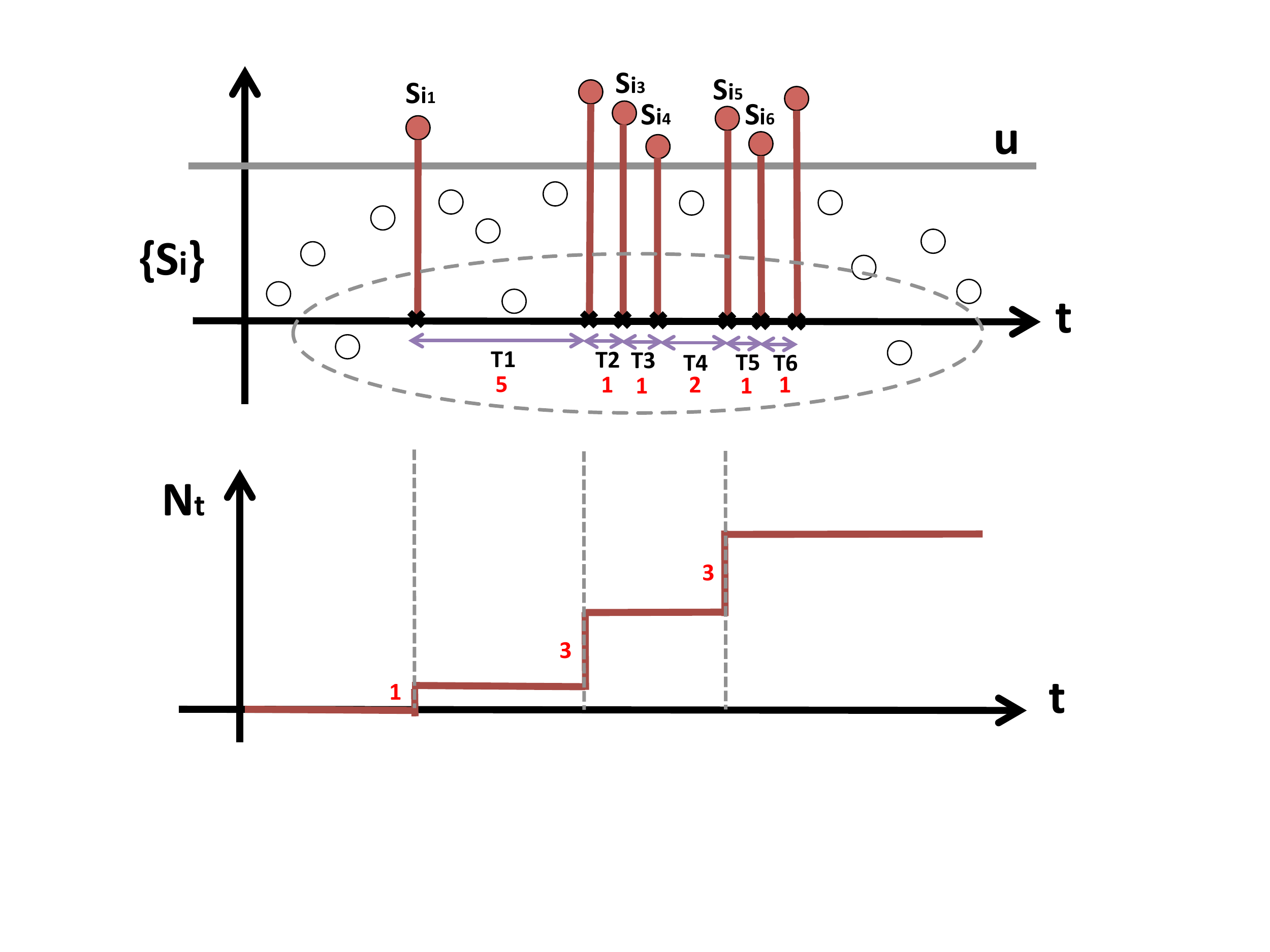}
    \vspace{-3mm}
    \caption{Stage III: Apply cutoff $u$ to $\mathbb{S}$ to obtain a compound Poisson process, and use the inter-exceedance time to estimate the extremal index $\theta$.}    
    \label{demo2}
  \vspace{-3mm}
  \end{figure}  
\end{itemize}

The overall algorithm is summarize in Algorithm~\ref{alg:Framwork}. The DTM algorithm can be applied when:
\begin{itemize}[leftmargin=*,nosep,nolistsep]
\item Sequence satisfies the so-called $\alpha$-mixing condition (\ref{condition}) , which is a moderate requirement. Sequences that satisfy the condition (\ref{condition}) include the $m$-th order Markov chain and the $m$-dependent sequence (i.e., two samples are independent if their indices are at least $m$ apart) \cite{Lehmann04}. Most machine learning algorithms with a finite memory of data will satisfy this requirement. 

\item Threshold $u$ should be chosen large enough so that the points exceed $u$ can be approximated as a Poisson process. Theorem 2.4.4 in \cite{Leadbetter83} states condition for the convergence of the exceedant points to a Poisson process. In practice, we choose $u$ as the .95 or .99 quantile of the data.

\item The number of samples $n$ should be large enough, so that we have enough samples exceeding a large $u$ for the estimation to be accurate, and also that asymptotic distribution of the maximum converges. In theory, the number of samples $n$ should be at least $O(\tau^2 e^{-\tau})$, where $\tau = - \log (1-\alpha)$, as a consequence of Theorem 2.4.2 in \cite{Leadbetter83}. In practice, when the number of exceedant sample has a moderate size, say, 10 to 100, the estimate for the threshold will still be accurate.
\end{itemize}

%
%

\begin{algorithm}[h]         
\renewcommand{\algorithmicrequire}{\textbf{Input:}}   
\renewcommand{\algorithmicensure}{\textbf{Output:}}  
\caption{Data-driven threshold machine (DTM).
}            
\label{alg:Framwork}                 
\begin{algorithmic}[1]                
\REQUIRE 
A sequence $\mathbb{S} = \{S_1, S_2, \dots, S_n\}$;  \\                    
Tail probability level $\alpha$;\\
Parameter $u$ to select exceedant sample.
\ENSURE ~~\              
threshold $x$ \\   
\COMMENT{{\bf Stage I: Boostrap sample}}
\STATE Bootstrap $\mathbb{S}$ to form~\iid~sample $\mathbb{S}^*$. \\
\STATE Select from $\mathbb{S}^*$ that exceeds $u$ and record their ``time'' (index) and heights $\{ (i_1, S^*_{i_1}), (i_2, S^*_{i_2}), \dots \}$; \\
\COMMENT{{\bf Stage II: Estimate parameters $\mu$, $\sigma$ and $\xi$}}
\STATE Use exceedant heights $\{S^*_{i_1}, S^*_{i_2}\ldots\}$ to estimate location $\hat{\mu}$, scale $\hat{\sigma}$, and shape $\hat{\xi}$ parameters that maximize the marked Poisson process likelihood function (\ref{likelihood});
\COMMENT{{\bf Stage III: Estimate extremal index $\theta$}}
\STATE Select from $\mathbb{S}$ that exceeds $u$ and record their ``time'' (index) and heights $\{ (i_1, S_{i_1}), (i_2, S_{i_2}), \dots \}$; \\
\STATE Use exceedant times $\{i_1, i_2, \ldots\}$ to estimate the extremal index $\hat{\theta}$
by maximizing the mixture model likelihood (\ref{likelihood_2});\\
\RETURN $x = \hat{C}^{-1}(-(1/\hat{\theta})\log(1-\alpha))$ where
\begin{equation}\label{C_hat}
\vspace{-2mm}
\hat{C} =
\left\{  
\begin{array}{ll}
\left[1 + \hat{\xi} \left(    \frac{x -\hat{\mu}}{\hat{\sigma}} \right) \right]^{- \frac{1}{\hat{\xi}}}, 
& \hat{\xi} \neq 0 \\
 \mbox{exp} \left\{  - \frac{x-\hat{\mu}}{\hat{\sigma}} \right\} & \hat{\xi} = 0.
\end{array}\right.
\vspace{-1mm}
\end{equation}
\end{algorithmic} 
\end{algorithm} 

%


\vspace{-4mm}
\section{Theoretical Derivation}
\vspace{-3mm}

DMT is based on the profound extreme value theory. We will show why DMT works. We first present the background of extreme value theory. Then we present how to estimate the three parameters $\mu$, $\sigma$, and $\xi$ for the so-called extreme value distributions using the heights of exceedant sample. Finally, we present how to estimate the extremal index $\theta$ using the time intervals between exceedant samples.

\vspace{-4mm}
\subsection{Parametrizing tail probability for extreme value}
\vspace{-3mm}

Essentially, the problem we want to solve is to estimate the tail probability of extreme values. Surprisingly, as we show in Theorem~\ref{theorem}, these extreme value distributions will follow specific parametric forms irrespective of the original distribution for $S_t$ and the dependence structure. Hence, our problem can be tackled by estimating the parameters of these parametric distributions. We will first describe the mixing condition needed for the theorem.  

%

\begin{definition}[Distributional mixing condition $\mathbf{D(u)}$] 
\vspace{-2mm} 
A stationary sequence $\{S_1,\dots,S_n\}$ is said to satisfy the distributional mixing condition, if  for any integers $ 1<i_1 < i_2< \dots < i_p< j_1 <j_2 < \dots < j_q<n$ for which  $j_1-i_p>l$, and for any real $u$
\begin{equation} \label{condition}
\begin{split}
&\left|  \mathbb{P}\left\{ S_{i_1}  \leq u, \dots, S_{i_p} \leq u, S_{j_1} \leq u, \dots, S_{j_q} \leq u  \right \} \right. \\
 & \quad - \mathbb{P}\left\{ S_{i_1}  \leq u, \dots, S_{i_p} \leq u \right\}  \cdot \\
 &\quad\quad \quad \left. \mathbb{P} \left\{ S_{j_1} \leq u, \dots, S_{j_q} \leq u \right\} \right|  \leq g(l),
\end{split}
\end{equation}
where $g(l)\rightarrow 0$ as $l\rightarrow \infty$.
\vspace{-3mm} 
\end{definition}
The distributional mixing condition is a mild condition, which ensures that the dependence between $S_i$ decay fast enough. It is satisfied in most learning scenarios. For instance, the~\iid~sequence, the order $M$ Markov chain, and the $M$ dependent sequence all satisfy (\ref{condition})~\cite{Lehmann04}. Most scan statistics satisfy (\ref{condition}) since the detection statistics are computed locally and any statistic computed over non-overlapping regions are mutually independent. With the above mixing condition, we can state the following fundamental extreme type theorem~\cite{Fisher28,Gnedenko43,Leadbetter83}
\begin{theorem}[Extreme type theorem.]\label{theorem}
\vspace{-2mm} 
Let $\{ S_1, \dots, S_n\}$ be a stationary process with marginal distribution $F$ and satisfying the distributional mixing condition (\ref{condition}). 
Let $\{S_1^*, S_2^*, \dots\}$ be another sequence of independent variables with the same marginal distribution $F$. Let 
\[M_n = \max_{1 \leq t \leq n} S_t, \quad \mbox{and} \quad M_n^* = \max_{1 \leq t \leq n } S_t^*.\] Then there exist a sequence of positive $\{a_n\}$ and positive $\{b_n\}$ such that
\begin{equation}
\begin{split}
& \mathbb{P}\left\{  \frac{M_n^*-b_n}{a_n} \leqslant x \right\} \xrightarrow{n\to \infty} G(x) \quad \mbox{and}\\
&  \mathbb{P}\left\{ \frac{M_n-b_n}{a_n} \leqslant x \right\} \xrightarrow{n\to \infty} [G(x)]^{\theta},
\end{split}
\end{equation}
where $\theta \in (0, 1]$ is the constant called the extremal index. Depending on the marginal distribution $F$, $G(x)$ is a member of the generalized extreme-value-distribution parameterized as
\begin{align} \label{gevd}
G(x) = 
  \begin{cases}
  \exp\{- \left[ 1 + \xi  \left( \frac{x-\mu}{\sigma}  \right)  \right]^{-\frac{1}{\xi}} \}, & \xi \neq 0; \\
  \exp\{ -  e^{   - \frac{x-\mu}{\sigma} } \},  & \xi =0,
  \end{cases}
\end{align}
defined over the set $\{x: 1 + \xi (x-\mu) / \sigma >0\}$, with location parameter $\mu$, scale parameter $\sigma>0$, and shape parameter $\xi$: $\xi>0$ corresponds to Fr\'echet distribution, $\xi <0$ corresponds to Weibull distribution, and $\xi = 0 $ corresponds to the Gumbel distribution. 
\vspace{-2mm} 
\end{theorem}


In plain words, this extreme type theorem states that for~\iid~sequences, the extreme value has to converge to one of three functional forms of the extreme value distribution, under the so-called ``distributional mixing condition''. For a dependence sequence, the asymptotic distribution can be constructed from an~\iid~sequence with the same marginal distribution and a so-called extremal index $\theta$, which is related to the local dependence of the sequence
$\{S_i\}$ at a extreme level. This theorem motivates our approach to estimate the tail probability in~\eq{key_problem}. Essentially, we will first construct an~\iid~sequence to estimate the parameters in $G(x)$, and then estimate the extremal index $\theta$ using the original dependent sequence.  

One may wonder how to find \(a_n\) and \(b_n\). In fact, it can be shown that $G( (x-b_n)/a_n)$ remains to be one of the three extreme value distributions just with different parameter values \cite{Leadbetter83}. Hence, we may estimate $\mathbb{P}\{M_n^* \leqslant x\}$ directly by estimating the three parameters of the extreme value distribution $G(x)$, without worrying about the specific form of \(a_n\) and \(b_n\). 

\vspace{-4mm}
\subsection{Learning parameters for $G(x)$} \label{sec:g}
\vspace{-3mm}

Thus, given the observed data \(\mathbb{S}=\{S_1, \dots, S_n\}\), which are dependent and stationary, we will first construct a sequence of \iid~data \(\mathbb{S}^*=\{S_1^*, \dots, S_n^*\}\) with the same marginal distribution to learn the extreme value distribution \(G(x)\). Thus, we will first bootstrap (or sample with replacement) from the original sequence $\mathbb{S}$ to generate the
new i.i.d. sequence $\mathbb{S}^*$. This sampling scheme preserves the marginal distribution $F(x)$ but breaks the local dependence in the original stationary sequence.

Next, given $\mathbb{S}^*$, we choose a high cutoff value $u$ to obtain the sequence of exceedant samples (as illustrated in Figure \ref{demo1}). In practice, $u$ is set to $.95$ or $.99$ quantile of the data. Let $n_u$ denote the random number of samples that exceed the cutoff $u$. Since this number depends on the choice of $u$, we use $u$ as the subscript. Let $\{i_1, i_2,\dots, i_{n_u}\}$ denote the index of these exceedant sample, and then 
$$\{S^*_{i_1}, S^*_{i_2}, \dots, S^*_{i_{n_u}}\}$$ 
are the selected exceedant points.

{\bf Marked Poisson process approximation.} To estimate parameters for $G(x)$, the key idea is a ``Poisson trick'': the normalized index of the exceedant sample can be approximated by a Poisson process, and the marks of the events will correspond to the heights of the exceedant sample. The precise statement can be found in Theorem 5.2.1 of \cite{Leadbetter83}. Below, we present a simple argument to show that the intensity of the process is related to the extreme value distribution $G(x)$. 

Since $S_t^*$ is an~\iid~sequence, we have that $\mathbb{P}\left\{M_n^*\leq u \right\} = F^n(u)$.
Alternatively, based on Theorem \ref{theorem}, we have that for large $n$, $\mathbb{P}\left\{M_n^*\leq u \right\}\approx G(x)$. By relating these two, and taking log on both sides, we obtain $n\log F(u) \approx \log G(x)$. Furthermore, for large $u$, $F(u)$ is close to 1, and $\log F(u) \approx -(1-F(u))$ using Taylor expansion.
Hence, we obtain 
$$1-F(u)\approx -(1/n) \log G(u),$$  
which means that for every data point, the probability to exceed the threshold $u$ is $-\log(G(u))/n$, a small number for large $u$. If we define a point process $N_n$ on the unit interval $(0, 1]$ consisting of events corresponding to normalized index of the exceedant sample, $\{i_1/n,\ldots,i_{n_u}/n\}$, then the point process converges to a Poisson process with intensity equal to $n(-\log(G(u))/n) = -\log G(u)$.

Further taking into account the heights of the exceedant sample, we can model the sequence of pairs, $\{ (\smallfrac{i_1}{n},S^*_{i_1}),\ldots,(\smallfrac{i_{n_u}}{n},S^*_{i_{n_u}} )\}$, as a {\it marked Poisson process} where the heights corresponds to the markers of the events. The intensity measure of the process for any set $\mathcal A=[\tau,t] \times (x, \infty)$ is hen given by
\begin{align}
\Lambda^*(\mathcal A) = 
\begin{cases}
(t-\tau) \left[  1+ \xi \left( \frac{x-\mu}{\sigma}  \right)  \right]^{-\frac{1}{\xi}}, & \xi \neq 0; \\
(t-\tau) e^{  - \frac{x-\mu}{\sigma} },  & \xi =0.
\end{cases}
\end{align}
Taking derivative, for any $t$ and $x \geqslant u$, we have the intensity function of the process given by
\begin{align}
\lambda^*(t, x) = 
\begin{cases}
 \sigma^{-1} \left[  1+ \xi \left( \frac{x-\mu}{\sigma}  \right)  \right]^{-\frac{1}{\xi} -1} & \xi \neq 0; \\
\sigma^{-1} e^{  - \frac{x-\mu}{\sigma} }  & \xi =0.
\end{cases}
\end{align}

{\bf Likelihood function.} Therefore, the likelihood function for $\Ecal=\{ (\smallfrac{i_1}{n},S^*_{i_1}),\ldots,(\smallfrac{i_{n_u}}{n},S^*_{i_{n_u}} )\}$ under the {\it marked Poisson process} model is given by 
\begin{equation}\label{likelihood}
\begin{split}
& \mathcal{L}(\mu, \sigma, \xi; \Ecal) = \exp \left\{ -\Lambda_0^* \right\} \prod_{k=1}^{{n_u}} \lambda^*(\smallfrac{i_k}{n}, S^*_{i_k}) \\
&  \propto 
  \left\{ 
  \begin{array}{ll}
  \exp \left\{ - \left[  1 + \xi \left( \frac{u-\mu}{\sigma}  \right) \right]^{-\frac{1}{\xi}}  \right\}\cdot \\
  \quad \prod_{k=1}^{{n_u}} \frac{1}{\sigma} \left[  1+ \xi \left( \frac{S^*_{i_k}-\mu}{\sigma} \right) \right]^{-\frac{1}{\xi}-1}, &
  \xi \neq 0 \\
  \exp  \left\{  - \exp \left\{  - \frac{u-\mu}{\sigma} \right\} \right\}
  \prod_{k=1}^{{n_u}} \frac{1}{\sigma} \cdot & \\
  \quad \exp \left\{  -  \frac{S^*_{i_k}-\mu}{\sigma} \right\}, & \xi = 0.
  \end{array}
  \right.
\end{split}
\end{equation}
where $\Lambda_0^*:=\Lambda^*((0,1]\times(u,\infty))$. 
From (\ref{likelihood}), we find that the likelihood function only depends on the heights of the exceedant sample. Once $u$ is fixed, the index of the exceedant sample does not change the likelihood function. 

Maximization of the likelihood function over the parameters does not lead to an analytical solution, but it can be done via standard optimization since only three variables are involved. Initialization is done with the method-of-moments, which relate the mean and variance of the exceedant sample to the three parameters-to-be-estimated, to avoid the discontinuity at $\xi = 0$. More details can be found in \cite{Coles03}. 

\vspace{-4mm}
\subsection{Learning extremal index $\theta$.}\label{sec:t}
\vspace{-3mm}

In this section, we focus on learning the extremal index $\theta$, which captures the dependence of the original sequence $\mathbb{S}$. Now we will apply the cutoff $u$ to $\mathbb{S}$ and obtain a new set of index $\{i_1,\ldots,i_{n_u}\}$, and the corresponding heights
$$
  \{S_{i_1}, \ldots, S_{i_{n_u}} \}. 
$$
We will use the inter-exceedance times to estimate $\theta$, based on a theory in \cite{Hsing88}. 

{\bf Compound Poisson process approximation.} Basically, when $\{S_1, \dots, S_n\}$ is stationary, the inter-exceedance times $\{i_1, i_2,\dots, i_{n_u}\}$ will converge to a { \it compound Poisson process}. A compound Poisson process is a continuous-time stochastic process with jumps. The jumps occur randomly according to a Poisson process, and the size of the jumps is also random according to a probability distribution (as illustrated in Figure~\ref{demo2}). 

Based on this theory, \cite{Ferro03, Suveges07}  give a more refined characterization. They proved that the limit distribution of the inter-exceedance times would be a mixture of an exponential distribution and a point mass on zero; the mixing proportion for the point mass will be equals to $\theta$. Intuitively,  when there is a dependency in the sequence, even in the extremal level, the data points tend to exhibit a clustering structure. If one data point reaches a high level, then the successive data tend to reach a high level as well. Hence, $\theta$ characterizes the clustering behaviors of the data at the extreme level and it can be interpreted as the inverse of the limiting mean cluster size. 

More specifically, let $T_k(u)$ denote the $k$-th inter-exceedance time, with $T_k(u) = i_{k+1}-i_k$, $k = 1, \dots, n_u-1$ (see Figure \ref{demo2} for an illustration). When $T_k(u)-1$ is nonzero, then the value of $T_k(u)-1$ can be interpreted as a distance between two adjacent clusters. Let $F(u)$ be the marginal probability that $S_i\leq u$. \cite{Ferro03,Suveges07} proved that when $n$ tends to infinity, the limiting distribution of the variable $(1-F(u)) (T(u)-1)$ converges a mixture distribution
\begin{equation}\label{compound_density}
\begin{split}
 & \mathbb{P} \left\{  (1-F(u))(T(u)-1) \in (t, t+dt)    \right\}\\
&= \left\{\begin{array}{ll}
1-\theta, & t = 0; \\
\theta^2 e^{-\theta t}, & t>0
\end{array}\right.
\end{split}
\end{equation}
This means that with probability $\theta$ the inter-exceedance time is an exponential variable with rate $\theta$, and otherwise it is of length zeros. Note that all zero observations of $T_k(u)-1$ will attribute to the point mass component of the likelihood.  


{\bf Likelihood function.} Using (\ref{compound_density}), we can write the likelihood function of the sequence of inter-exceedance time, $\{T_1(u)-1,\ldots, T_{n_u-1}(u)-1\}$, from which we can estimate $\theta$ 
\begin{equation} \label{likelihood_2}
\begin{split}
& \mathcal{L}( \theta; S_1, \dots, S_n) 
 = (1-\theta) ^{  (n_u - n_c - 1)  } \theta^{2 n_c  } \\
 & \quad  \mbox{exp} \left\{   -\theta \sum_{i=1}^{n_u-1}
(1-F(u))(T_i(u)-1)
   \right\}, 
   \end{split}
\end{equation}
where $n_c =\sum_{i=1}^{n_u-1} \mathbb{I}\{(T_i-1) \neq 0\}$ corresponds to the number of non-zero inter-exceedance times, and $(1-F(u))$ can be replaced by its estimate $n_u/n$. A closed form expression for the maximum likelihood estimator $\hat{\theta}$ can be easily derived: 
\[
\hat{\theta} = 1 - \frac{n_u - n_c - 1}{
2n_c - \sum_{j=1}^{n_u - 1}(1-F(u))(T_j(u) - 1)}.
\]
Thus, this estimator of $\hat \theta$ together with the estimators for $\hat \mu$, $\hat \sigma$ and $\hat \xi$ from Section~\ref{sec:g} completes the major work of our data-drive threshold machine. Last, we set  $G(x; \hat \mu, \hat \sigma, \hat xi)^{\hat \theta} = \alpha$ and solve for $x$ and obtain $x = {\hat C}^{-1} (-(1/{\hat \theta})\log(1-\alpha))$ as used in Algorithm~\ref{alg:Framwork}. 

\vspace{-4mm}
\section{Numerical Examples}
\vspace{-3mm}

We will conduct two set of examples in this section investigating the accuracy of tail probability estimation and applying our method to a few machine learning problems.

\vspace{-4mm}
\subsection{Accuracy of tail probability modeling.}
\vspace{-3mm}

We study the accuracy of the DTM in estimating of $\mathbb{P}\{ \max_{1 \leq i \leq n} S_i \leq x \}$ by comparing with the simulation results. First, we generate a total number of $L$ sequences each with sample size $n$. For each sequence, we record the maximal value. Then for the $L$ sequences, we will have $L$ such maximums. In this way, we can get the empirical distribution for $\mathbb{P}\{ \max_{1 \leq i \leq n} S_i \leq x \}$. If $L$ is a large number, we can regard this empirical distribution as the true distribution. On the other hand, we apply our algorithm to just one sequence of data with sample size $n$, and select the data points exceeding the predetermined $u$ to fit the model. Then substitute the estimated parameters into the parametric form to get the approximation. Note that our algorithm only uses $1/L$ of the amount the data compared to simulation.

\begin{figure}[h!]
\vspace{-3mm}
  \centering               
  \includegraphics[width=0.70 \linewidth]{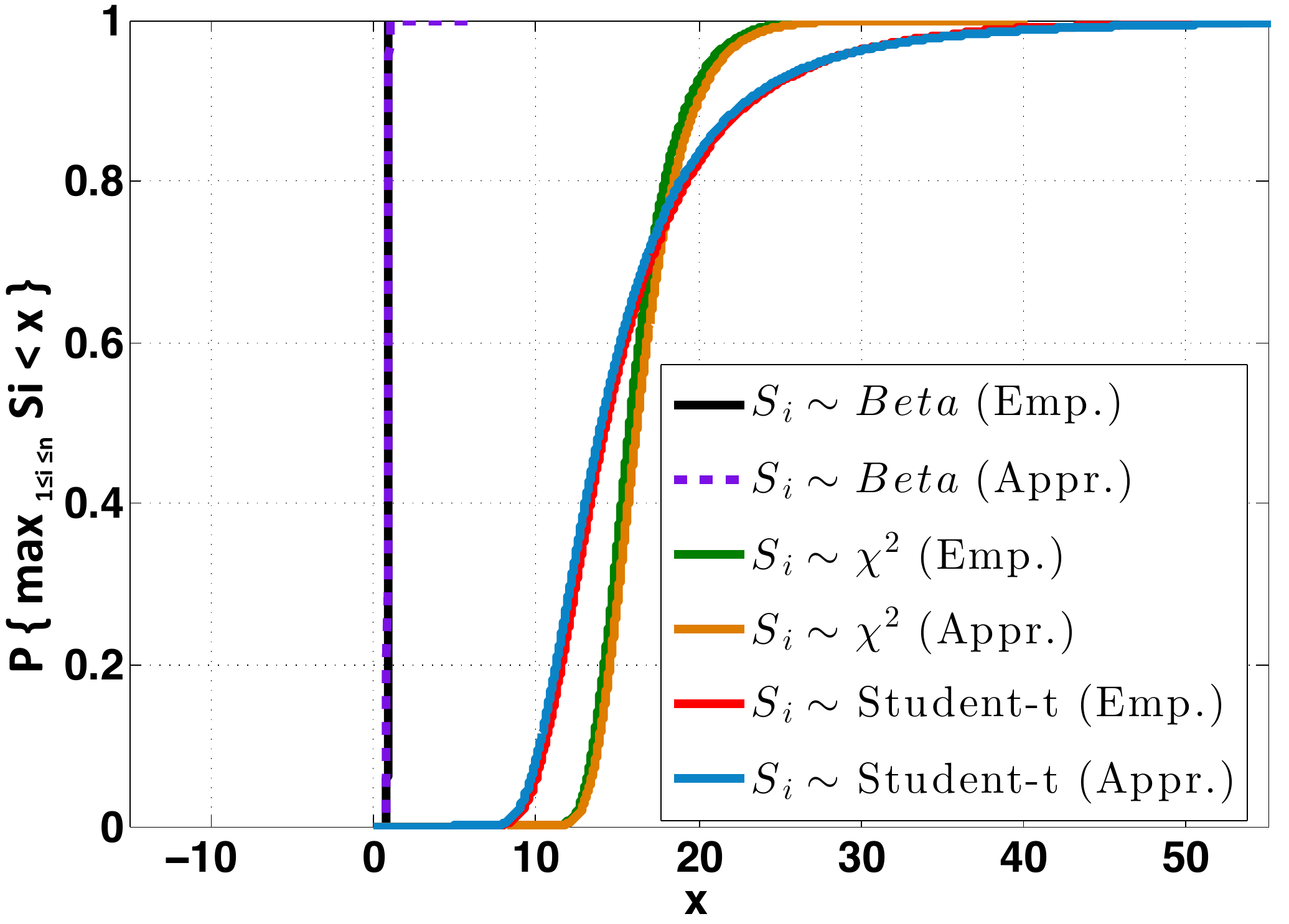}
  \vspace{-3mm}
  \caption{Adaptive to data distribution: comparison of empirical and approximated $\mathbb{P}\{ \max_{ 1\leq i \leq n} S_i\}$.}    
  \label{tail}
\end{figure}

{\bf Adaptive to data distribution.} We arbitrarily select one distribution from the three types of distributions, with exponentially decaying tails, heavy tails, and short tails respectively, and show DTM is agnostic to the underlying distribution.
Specifically, we consider the following cases: (1) $S_i \sim Beta(2, 5)$, which is short-tailed and the random variables is upper bounded by 1; (2) $S_i \sim \chi^2$ with degree 1, which has exponentially decaying tail, and (3) $S_i \sim $ Student-$t$ with degree 4, which is a heavy tail distribution.
Let $n=10^4$, $L=10^4$, and $u$ be the $.99$ quantile of the data. The comparison results of the empirical and the approximated distributions for $\mathbb{P}\{ \max_{ 1\leq i \leq n} S_i\}$, under the three cases are demonstrated in Figure {\ref{tail}}. Note that our algorithm only uses $10^{-4}$ of the amount the data compared to simulation, but get almost the same results.

In this example, we consider the following cases: (1) $S_i \sim Beta(2, 5)$, which is short-tailed and the random variables is upper bounded by 1; (2) $S_i \sim \chi^2$ with degree 1, which has exponentially decaying tail, and (3) $S_i \sim $ Student-$t$ with degree 4, which is a heavy tail distribution.
Let $n=10^4$, $L=10^4$, and $u$ to be the $.99$ quantile of the data. The results are demonstrated in Figure {\ref{tail}}. Note that our algorithm only uses $10^{-4}$ of the amount the data compared to simulation, but get almost the same results. Moreover, our algorithm does not need to know any prior knowledge about the tail of the underlying distribution. That is, we don't need to know whether the data $\{S_i\}$ are a heavy tail, short tail or exponentially decaying tails. The algorithm can adaptively and accurately learn this information from the data.

{\bf Adaptive to dependence.}
We study the accuracy of the DTM on stationary sequences $\{S_t\}$ with local dependence. Specifically, we consider the following random sequences
\[S_t = e^{-1/m} S_{t-1}+ \sqrt{1- e^{-2/m}} Z_t\]
where $\{Z_t\}$ are independent standard normal variables. For this sequence, it has such properties that $\{S_t\}$ is a Gaussian process with $\mathbb{E}[S_t]=0$, and $\mbox{Cov} (S_t, S_{t'})= \mbox{exp}(-|t'-t|/m)$. By adjusting $m$, we can control the strength of local dependence. If $m=0$, $\{S_t\}$ is an \iid~sequence. Increasing $m$ will enhance the local dependence. 

\begin{figure}[h]
\vspace{-3mm}
  \centering               
  \includegraphics[width=0.70 \linewidth]{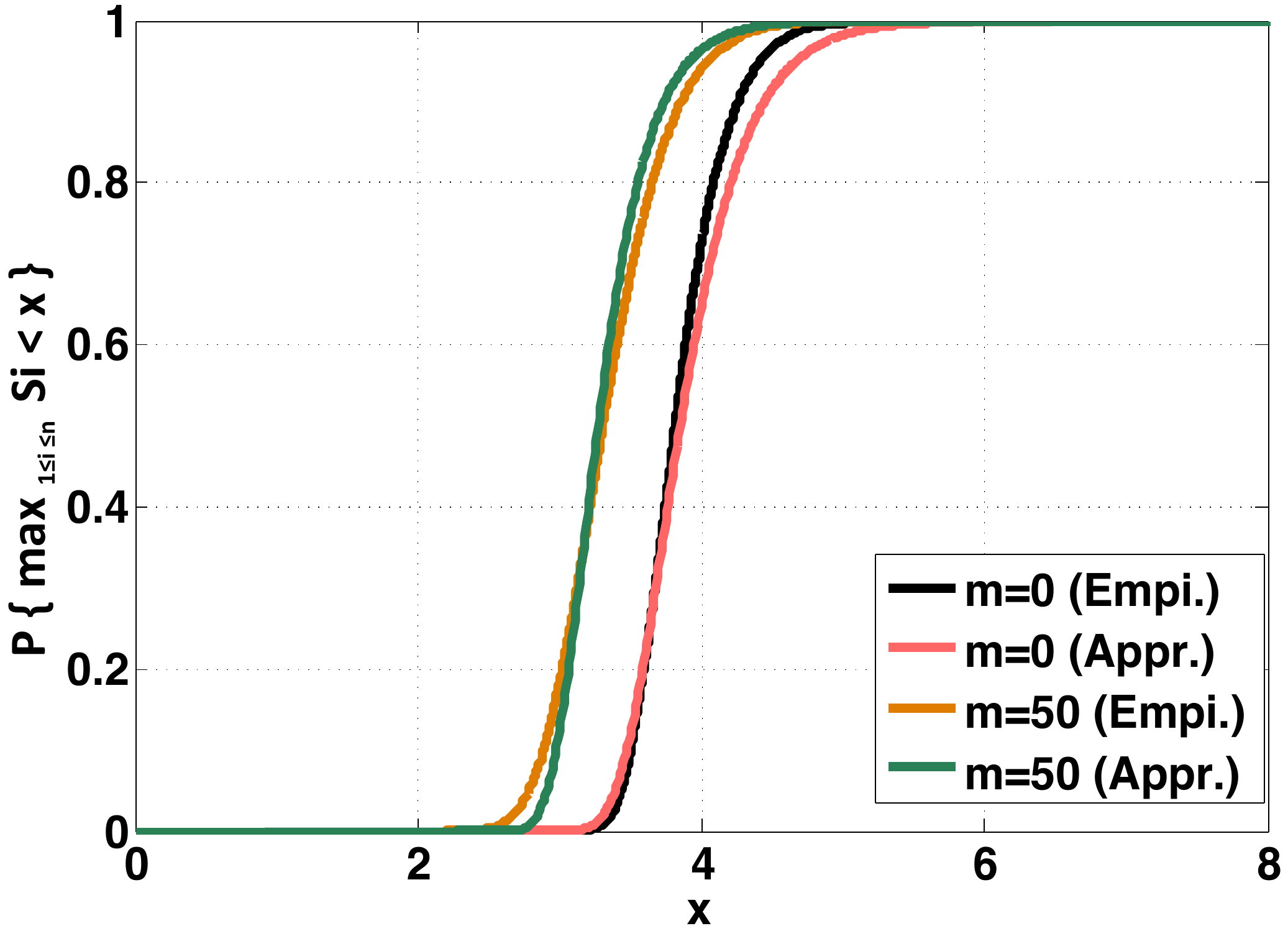}
  \vspace{-3mm}
  \caption{Adaptive to dependence: comparison of empirical and approximated $\mathbb{P}\{ \max_{ 1\leq i \leq n} S_i\}$.}    
  \label{tail_m}
 \vspace{-3mm}
\end{figure}

Consider $m=0$ and $m=50$, respectively. The values of $n, L$ are the same as previous examples. The comparison results of the empirical and the approximated distributions for $\mathbb{P}\{ \max_{ 1\leq i \leq n} S_i\}$ are displayed in Figure \ref{tail_m}. Our algorithm shows consistent results with simulation. The estimated extremal index are $\hat{\theta}=1.000$ and $\hat{\theta}=0.246$ in these two examples. We know that $\theta=1$ corresponds to the independent sequence, and increase the local dependence of the random process, $\theta$ will be more close to 0. This means, our algorithm can accurately learn the distribution $\mathbb{P}\{ \max_{ 1\leq i \leq n} S_i\}$ for dependent sequence. Moreover, we don't need to know beforehand whether $\{S_i\}$ are independent or dependent. The algorithm can adaptively and accurately learn this information from data, reflected in the estimated values of $\theta$. 

\vspace{-4mm}
\subsection{Application to choice of threshold}
\vspace{-3mm}

{\bf Scan statistics over graph.} 
We consider the problem of community detection which has been studied in \cite{randomgraph14}. The problem is cast into detecting a dense subgraph in a random graph. The null hypothesis is that the random graph is an Erdos-Renyi Graph, i.e., edges between nodes are \iid~Bernoulli random variables with probability $p_0$ being one. Alternatively, there is a subgraph such that the edges are formed with higher probability $p_1 > p_0$. Let $W_{ij}$ denote the adjacency matrix of the random. The scan test  detects a community when the statistic $\max_{\mathcal G} \sum_{(i, j) \in {\mathcal G}} W_{ij} > x$, where $\mathcal{G}$ denotes a subgraph contains the community and $x$ is the threshold. Let $N$ be the number of nodes. If we assume the size of the community is $k$, there are $N \choose k$ such $\mathcal{G}$. Since $N \choose k$ is usually a very large number, we randomly pick $n$ possible $\mathcal{G}$ when forming the scan statistics.

We consider the case where $N = 100$, $p_0 = 0.1$, $k = 10$, and $n = 5000$. The Monte Carlo results are obtained from 100 repetitions of the experiments. As shown in Table \ref{table1}, the threshold obtained via DTM is consistent with and higher that obtained from Monte Carlo simulation (in fact, the Monte Carlo results, in this case, are obtained from a relatively small number of repetitions; hence the estimated thresholds from Monte Carlo tend to be small).

\begin{table}[h]
\vspace{-0.1in}
\caption{Scan over random graph, threshold obtained via Monte Carlo simulation versus DTM. }
\begin{center}
\begin{tabular}{c|cccc}
\hline
$\alpha$ & 0.1 & 0.05 & 0.03 & 0.01  \\\hline
Monte Carlo & 12.00 & 12.00 & 13.00 & 13.00\\
DTM & 13.71 & 14.50 & 14.64 & 14.55\\\hline
\end{tabular}
\end{center}
\label{table1}
\vspace{-0.1in}
\end{table}

{\bf Change-point using MMD statistic.} We show that DTM can aid change-point detection in the online setting. In this example, the objective is to detect the activity changes over the network by monitoring the adjacency matrix $W$. Still let $N=100$, and the observations are a snapshot of a realization of the adjacency matrix with dimension 100 by 100. Let $p_0=0.3$ before the change-point and $p_1=0.4$ after the change-point. The true change-point occurs at time $4000$. We introduce the {\it maximum mean discrepancy} (MMD) as the detection statistic and use the online sliding window search scheme to monitor potential changes. The experiment setting is the same as \cite{Li15}. We set the block size $B=50$ and only use one block. Every time, MMD is formed by the to-be-test data $X$ within the sliding block with  and the reference data $Y$ with the same size by
$\text{MMD}^2[X, Y] = \frac{1}{B(B-1)} \sum_{i, j=1, i \neq j}^B h(x_i, x_j, y_i, y_j),$ 
where $h(x_i, x_j, y_i, y_j)=k(x_i, x_j)+k(y_i, y_j)-k(x_i, y_j)-k(x_j, y_i)$ and $k(\cdot)$ is the kernel function (we are using Gaussian kernel). It is well-known that MMD is a nonparametric statistic and the asymptotic distribution is an infinite summation of weighted chi-squares, which has no closed-form. When applying MMD to the online change-point detection, we need to characterize the tail probability of the maximal value of MMD over time, and from which get the threshold indicating when to stop the algorithm and make a decision that there is a change-point.  
\begin{figure}[h]
  \vspace{-3mm}
  \centering               
  \includegraphics[width=0.75 \linewidth]{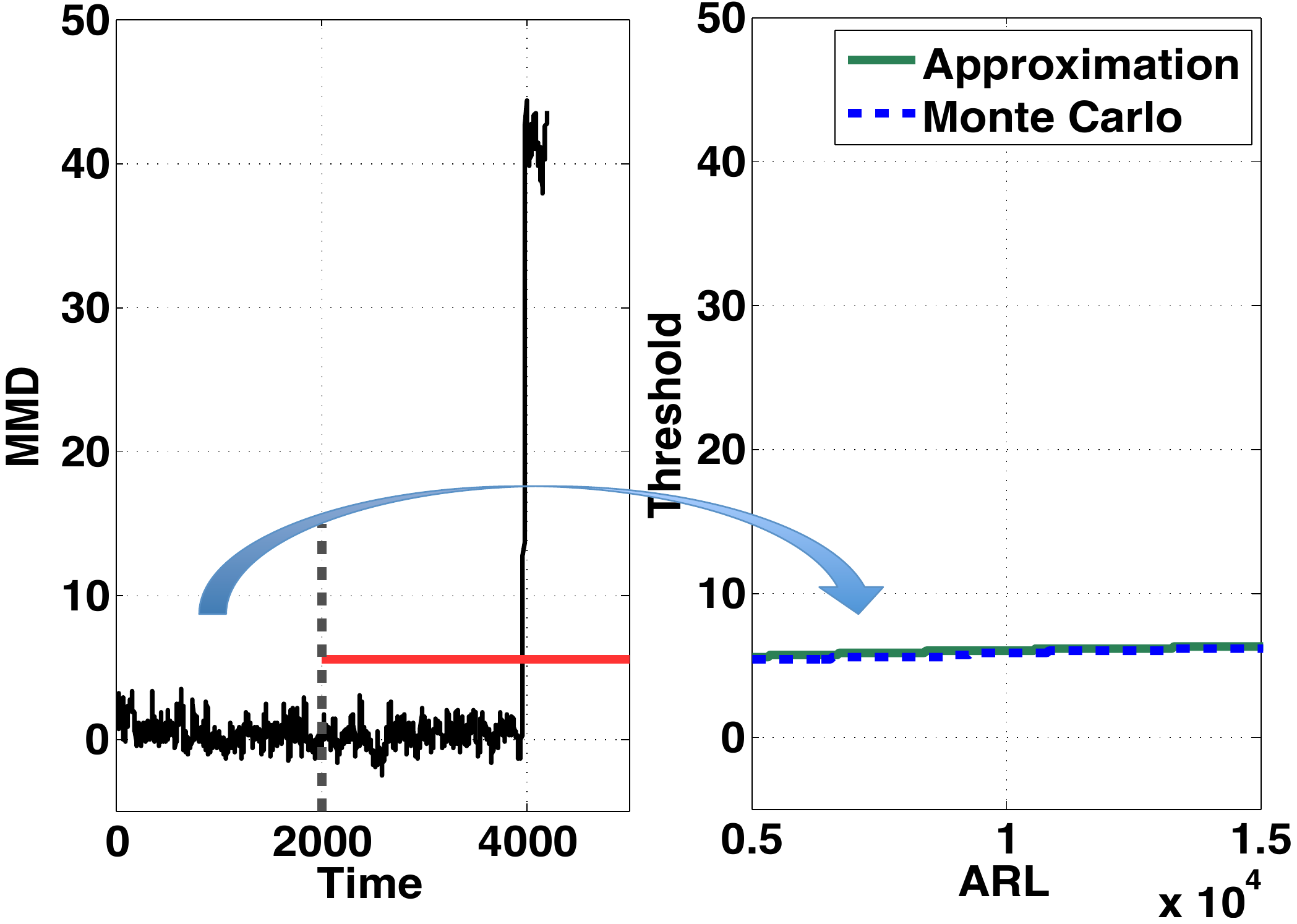}
  \vspace{-3mm}
  \caption{Change-point using MMD statistic.}    
  \label{MMD}
 \vspace{-3mm}
\end{figure}

DTM provides a cheap and accurate approach to getting the threshold. As shown in Figure \ref{MMD}. We first apply the detection algorithm on the raw data (realizations of adjacency matrix), and get a sequence of $\{ \text{MMD}_t\}$, which are our $\{S_t\}$ in DTM. Given $n=2000$ samples of $\{ {\text{MMD} }_{t}\}$, the estimated parameters are $(\hat{\sigma}, \hat{\xi}, \hat{\mu}, \hat{\theta})=(0.647, 0.000, 5.717, 0.306)$. Let ARL range from $.5\times 10^4$ to $1.5 \times 10^4$. The relation of threshold $x$ and ARL can be computed by $\mathbb{P}_0\{\max_{1\leqslant i \leqslant n} S_i > x\} = 1-e^{-n/{\rm ARL}}$. We compare this result with Monte Carlo, as shown on the right panel of Figure \ref{MMD}, and get the consistent results. For $\text{ARL}=.5\times 10^4$, we get the approximated threshold $x=5.54$ (this value for Monte Carlo is 5.35). We mark this threshold as the red line (as shown in the left panel of Figure \ref{MMD}) and sequentially monitor the change-point. It shows that the threshold successfully detects the change-point occurring at time 4000. 

Note that DTM works on the detection statistics directly, whereas for Monte Carlo method one needs to generate realizations of the adjacency matrix and form the MMD to get an approximation because the closed-form of MMD is unknown, which would be rather computationally expensive.

{\bf Max K-armed bandit.} 
In the max K-armed bandit (or called extreme bandit) setting, the objective is to find the best arm defined as the arm with the heaviest tail. 
Consider the following setting and algorithm. There are $K$ arms, each with underlying distribution $P_k$. At any time $t$, the the policy only choose one arm $i$ to pull and get one observation of the reward defined as $R_{i, s}, s= 1, 2, \dots$. Let $N_i(t)$ denote the number of times that arm $i$ has been sampled up to time $t$.
For any arm $i$, given the $N_i(t)$ observations till now, we define the upper confidence bound (UCB) $x_i^{\text{up}}$ to be:
$P\left\{  \max_{ 1\leq s \leq N_i(t)} R_{i,s} > x^{\text{up}}_{i}   \right\} =\delta$ 
and define the lower confidence bound (LCB) $x_i^{\text{low}}$ to be:
$P\left\{  \max_{ 1\leq s \leq N_i(t)} R_{i,s} < x^{\text{low}}_{i}    \right\} =\delta$,
 where $\delta$ is the confidence parameter. UCB and LCB play a crucial role in identifying the best arms in many online algorithm, such as Action Elimination (AE) algorithm, Upper Confidence Bound (UCB) algorithm, and LUCB algorithm \cite{Jamieson14}. For example, in the UCB algorithm, every time we choose to pull the arm with the highest UCB to get the reward. And stop the algorithm whenever the LCB for the best arm till now is higher than the UCBs for any other arms.
 
\begin{figure}[h]
\vspace{-3mm}
  \centering               
  \includegraphics[width=0.75 \linewidth]{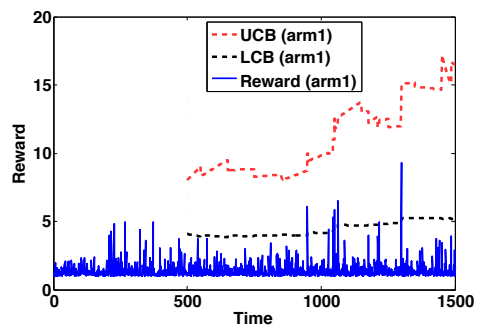}
  \vspace{-3mm}
  \caption{Adaptive UCB and LCB for rewards.}    
  \label{reward}
 \vspace{-4mm}
\end{figure}

 Our DTM provides a data-driven approach to estimating the UCB and LCB based on the real observed rewards to date, and can adaptively update the estimations given new observations. As a illustration, we consider the following example. We let $K=2$, and consider the Pareto distribution $P_k(x) =1-x^{-\alpha_k}$, where $\alpha = [3.5, 4.0]$. Fix $\delta=.005$. For the first example, rewards are \iid~generated from the underlying distributions. We first sample the two arms a fixed number of times. In the experiment, this number is 500. Then we can adopt DTM to estimate (LCB, UCB) from the history observations. The estimation results are (4.086, 8.001) for arm one, and (3.914, 7.172) for arm two. Then if we use the UCB algorithm, we will pull arm one in the next step. We find that the algorithm will stick to pulling arm one, which is the best arm. And we also demonstrate the adaptive UCB and LCB for arm one in Figure \ref{reward}. 

\begin{figure}[h]
\vspace{-3mm}
  \centering               
  \includegraphics[width=0.75 \linewidth]{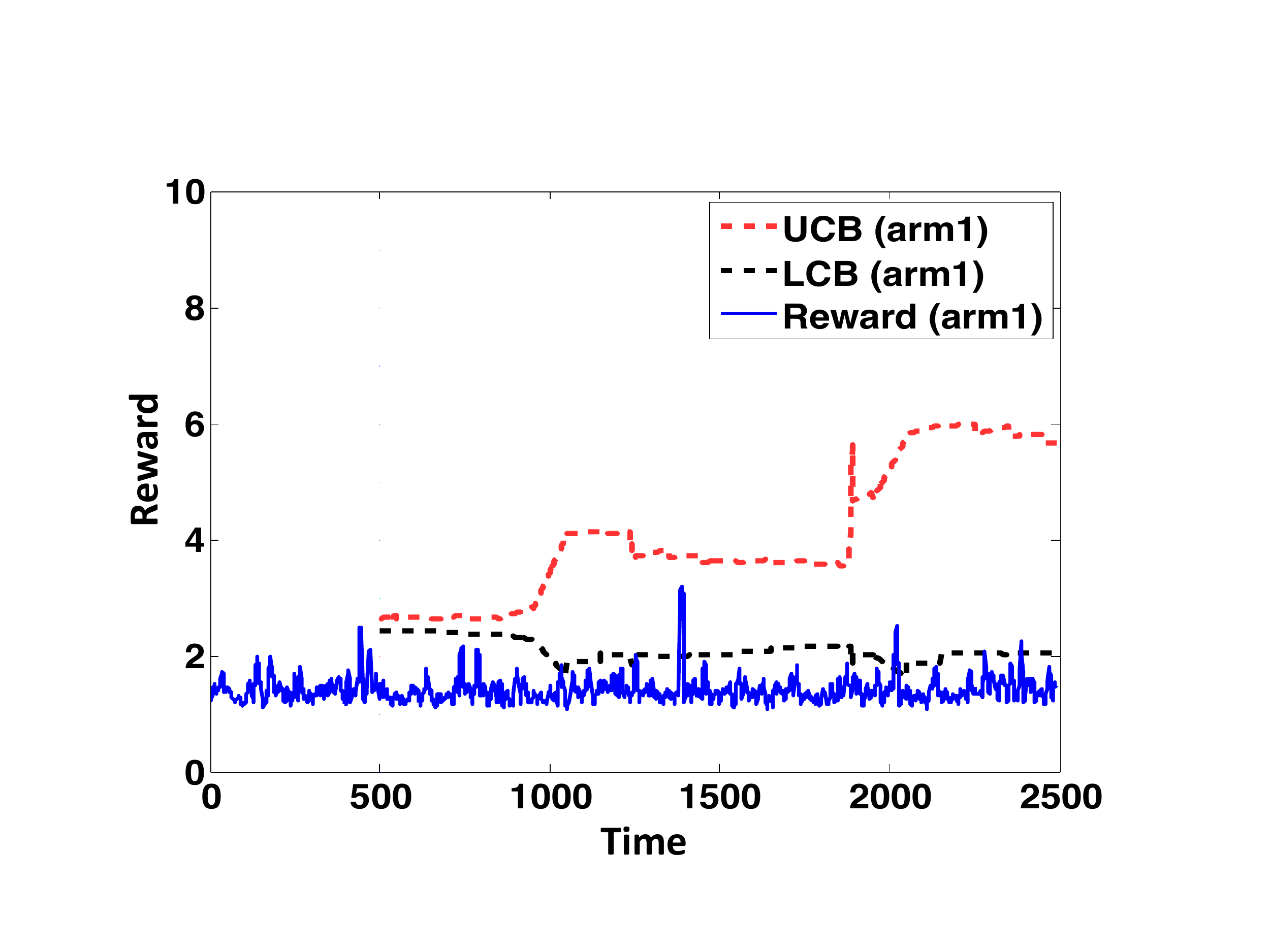}
  \vspace{-3.5mm}
  \caption{Adaptive UCB and LCB for dependent rewards.}    
  \label{reward_dep}
 \vspace{-2mm}
\end{figure}

Next, we consider the case where the observed rewards are stationary and temporal dependent for each arm. The setting is the same, however, the rewards for each arm is generated as a moving average of the first example. The introduced moving window would induce the dependence of the observations. Set the window to be 10. After 500 observations, the initial estimation for (LCB, UCB) is (2.439, 2.638) for arm one, and (1.614, 2.085) for arm two. We also demonstrate the adaptive UCB and LCB for arm one if we continue to pull it, as displayed  in Figure \ref{reward_dep}. 

%

\vspace{-4mm}
\section{Conclusion}
\vspace{-3mm}

We present a novel distribution-free approach, the
data-driven threshold machine (DTM), to choose threshold such that the extreme values are bounded by a pre-specified level. DTM only requires one sample path for a reliable estimate of the threshold. Numerical examples demonstrate the robustness of the method. As of future work, our approach can be extended using the general Khintchine's theorem to find a lower threshold for the lower tail.

\small
\bibliographystyle{alpha}
\bibliography{bibfile}

\begin{thebibliography}{SRMB11}

\bibitem[ACV14]{randomgraph14}
E.~Arias-Castro and N.~Verzelen.
\newblock Community detection in dense random networks.
\newblock {\em Ann. Statist.}, 2(3):940--969, 2014.

\bibitem[BC10]{threshold10}
J.~Broadwater and R.~Chellappa.
\newblock Adaptive threshold estimation via extreme value theory.
\newblock {\em {IEEE} Trans. Signal Process.}, 58(2):490--500, 2010.

\bibitem[CHT09]{Novelty09}
D.~A. Clifton, S.~Hugueny, and L.~Tarassenko.
\newblock Novelty detection with multivariate extreme value theory, part i: A
  numerical approach to multimodal estimation.
\newblock In {\em IEEE Workshop on Machine learning for signal processing
  (MLSP)}, 2009.

\bibitem[Col03]{Coles03}
S~Coles.
\newblock An introduction to statistical modelling of extreme values, software
  package “ismev” written in r language, 2003.

\bibitem[CS05]{kArm05}
V.~Circirello and S.~F. Smith.
\newblock The max k-armed bandit: A new model of exploration applied to search
  heuristic selection.
\newblock In {\em Proc. of Twentieth National Conference on Artificial
  Intelligence}, 2005.

\bibitem[CV14]{extremeBandits14}
A.~Carpentier and M.~Valko.
\newblock Extreme bandits.
\newblock In {\em Advances in Neural Information Processing Systems 27}, pages
  1089--1097, 2014.

\bibitem[FS03]{Ferro03}
C.~A.~T. Ferro and J.~Segers.
\newblock Inference for clusters of extreme values.
\newblock {\em J. Royal Statist. Soc. Ser. B}, 65(2):545--556, 2003.

\bibitem[FT28]{Fisher28}
R.A. Fisher and L.H.C. Tippett.
\newblock Limiting forms of the frequency distribution of the largest or
  smallest member of a sample.
\newblock In {\em Mathematical Proceedings of the Cambridge Philosophical
  Society}, volume~24, pages 180--190. Cambridge Univ Press, 1928.

\bibitem[GK06]{EVTRisk06}
M.~Gilli and E.~Kellezi.
\newblock An application of extreme value theory for measuring financial risk.
\newblock {\em Computational Economics}, 27(2):207--228, 2006.

\bibitem[Gne43]{Gnedenko43}
B.~Gnedenko.
\newblock Sur la distribution limite du terme maximum d'une serie aleatoire.
\newblock {\em Annals of mathematics}, pages 423--453, 1943.

\bibitem[GPW09]{Scan09}
J.~Glaz, V.~Pozdnyakov, and S.~Wallenstein.
\newblock {\em Scan statistics}.
\newblock Springer, 2009.

\bibitem[HHL88]{Hsing88}
T.~Hsing, J.~Husler, and M.~R. Leadbetter.
\newblock On the exceedance point process for a stationary sequence.
\newblock {\em Probab. Th. Rel. Fields}, 78(97), 1988.

\bibitem[JMNB14]{UCB2014}
K.~Jamieson, M.~Malloy, R.~Nowak, and S.~Bubeck.
\newblock lil' ucb: An optimal exploration algorithm for multi-armed bandits.
\newblock {\em JMLR: Workshop and Conference Proceedings}, 35:1--17, 2014.

\bibitem[JN14]{Jamieson14}
K.~Jamieson and R.~Nowak.
\newblock Best-arm identification algorithms for multi-armed bandits in the
  fixed confidence setting.
\newblock In {\em Information Sciences and Systems (CISS), 2014 48th Annual
  Conference on}, pages 1--6. IEEE, 2014.

\bibitem[Leh04]{Lehmann04}
E.L. Lehmann.
\newblock {\em Elements of large-sample theory}.
\newblock Springer, 2004.

\bibitem[LLR83]{Leadbetter83}
M.~R. Leadbetter, G.~Lindgren, and H.~Rootzen.
\newblock {\em Extremes and related properties of random sequences and
  processes}.
\newblock Springer, 1983.

\bibitem[LXDS15]{Li15}
S.~Li, Y.~Xie, H.~Dai, and L.~Song.
\newblock M-statistic for kernel change-point detection.
\newblock In {\em Advance in Neural Information Processing Systems}, 2015.

\bibitem[Roc12]{Rocco12}
M.~Rocco.
\newblock Extreme value for finance: A survey.
\newblock {\em Journal of Economic Surveys}, 2012.

\bibitem[Sie85]{Siegmund1985}
D.~O. Siegmund.
\newblock {\em Sequential Analysis: Tests and Confidence Intervals}.
\newblock Springer Series in Statistics. Springer, Aug. 1985.

\bibitem[Smi89]{Smith89}
R.~Smith.
\newblock Extreme value analysis of environmental time series: An application
  to trend detection in ground-level ozone.
\newblock {\em Statistical Science}, 4(4):367--377, 1989.

\bibitem[SRMB10]{ECCV10}
W.~Scheirer, A.~Rocha, R.~Micheals, and T.~Boult.
\newblock Robust fusion: Extreme value theory for recognition score
  normalization.
\newblock {\em ECCV}, 2010.

\bibitem[SRMB11]{PAMI11}
W.~Scheirer, A.~Rocha, R.~J. Micheals, and T.~E. Boult.
\newblock Meta-recognition: The theory and practice of recognition score
  analysis.
\newblock {\em {IEEE} Trans. Pattern Anal. Mach. Intell.}, 8(33):1689--1695,
  2011.

\bibitem[Suv07]{Suveges07}
M.~Suveges.
\newblock Likelihood estimation of the extremal index.
\newblock {\em Extremes}, 10:41--55, 2007.

\bibitem[SV95]{SiegmundVenkatraman1995}
D.~Siegmund and E.~S. Venkatraman.
\newblock Using the generalized likelihood ratio statistic for sequential
  detection of a change-point.
\newblock {\em Ann. Statist.}, 23(1):255 -- 271, 1995.

\bibitem[SYX13]{Shao13}
W.~Shao, W.~Yang, and G.-S. Xia.
\newblock Extreme value theory-based calibration for the fusion of multiple
  features in high-resolution satellite scene classification.
\newblock {\em Int. J. Remote Sensing}, pages 8588--8602, 2013.

\bibitem[XS13]{annals2013}
Y.~Xie and D.~Siegmund.
\newblock Sequential multi-sensor change-point detection.
\newblock {\em Annals of Statistics}, 41(2):670--692, 2013.

\end{thebibliography}

\end{document}